\documentclass{article}
\usepackage{amsmath}
\usepackage{float}  
\usepackage{graphicx} 

\title{Natural Language Processing and Multimodal Stock Price Prediction}
\author{Kevin Taylor, Jerry Ng}
\date{July 2023}
\begin{document}

\maketitle

\begin{abstract}
In the realm of financial decision-making, predicting stock prices is pivotal. Artificial intelligence techniques such as long short-term memory networks (LSTMs), support-vector machines (SVMs), and natural language processing (NLP) models are commonly employed to predict said prices. This paper utilizes stock percentage change as training data, in contrast to the traditional use of raw currency values, with a focus on analyzing publicly released news articles. The choice of percentage change aims to provide models with context regarding the significance of price fluctuations and overall price change impact on a given stock. The study employs specialized BERT natural language processing models to predict stock price trends, with a particular emphasis on various data modalities. The results showcase the capabilities of such strategies with a small natural language processing model to accurately predict overall stock trends, and highlight the effectiveness of certain data features and sector-specific data.
\end{abstract}

\section{Introduction}

Stock price prediction is a crucial part of financial decision making. Previous works in this field tended to use artificial intelligence technologies such as long short-term memory networks (LSTMs)\cite{9257950, pmlr-v95-li18c, 9526491,7364089}, support-vector machines (SVMs)\cite{doi:10.1080/03081070601068595, 6706743, 6703096}, and natural language processing (NLP) models\cite{8806182,COLASANTO2022341,8848203, 9759178}. These techniques are relevant to the task of stock market prediction. LSTM models are well-suited for time series predictions, which are useful in stock price prediction because the price of a stock can follow certain trends and cycles over time. SVM models are adept at handling high-dimension and nonlinear data. Since most stocks are affected by multiple factors(dimensions) and do not have linear price trends, SVM networks are a good candidate for prediction. Finally, NLP models are useful because they can analyze textual sentiment, giving them the ability to gauge public opinion about a stock based on news headlines, and such articles can impact whether a stock rises or falls in a market.

This paper is not the first on the prediction of stock prices using natural language processing, but it does touch upon a new technique: the use of percentage change as training data. Most models that aim to predict the stock market use raw currency values\cite{8920761}, and very few papers have incorporated relative price increase and decrease via the use of percent change into study, indicating that this topic is prominent for research.
This work aims to understand how the behavior of stocks -- such as their price direction -- may be accurately predicted, and gain a deeper understanding of how the market behaves. This goal is accomplished by leveraging sentiment analysis on publicly released news articles, and training a model using stock value percent change instead of raw price values. The reasoning behind using percent change instead of raw dollar price is to help models make more accurate decisions by marking changes in the context of overall company stock price. A stock worth \$50 dropping \$5 in value is a lot more significant than a stock worth \$6,000 dropping \$5. Arming the model with stock percent change allows it to see the scope of impact that the price change has. The models should be able to determine that \$5 isn't as significant to a stock at \$6000(0.083\% change), aiding in their predictive accuracy.
\section{Methodology}

For our primary dataset, we searched the web for news articles on a given set of dates for each stock in the S\&P 500. The total number of data points is approximately 8,000, with all collected during the 2022-2023 window. Our secondary dataset was closer to 700 points, and focused specifically on the area of physical and virtual technology companies like Google, Amazon, and Tesla.
We created the tech subclass in order to better study how predictive techniques might be specialized to certain industries or fields.
\begin{figure}[H]
    \centering
    \includegraphics[width=1\linewidth]{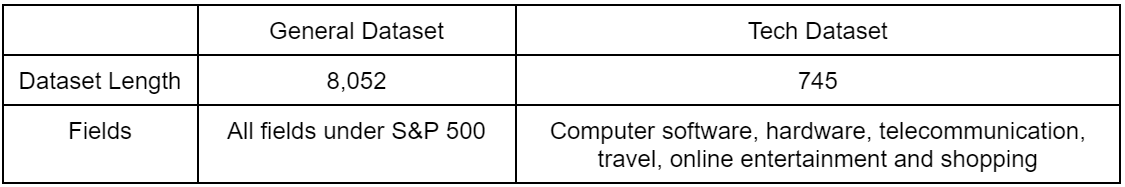}
    \caption{Dataset Table}
    \label{fig:Dataset Table}
\end{figure}
We tested our model by modifying the modalities it was able to access. For each news article found, a data point was added to the collection containing key information such as:
\begin{itemize}
    \item The article headline
    \item The article source
    \item The date
    \item The company and stock symbol
    \item The price before, after, and the percent change from the market opening to closing on a selected date
\end{itemize}
  The versions/modalities of data given to the model are as follows:
\begin{itemize}
  \item 1st version: Article Headlines, Article Sources, Company Name
  \item 2nd version: Article Headlines, Company Name
  \item 3rd version: Article Headlines, Article Sources
  \item 4th version: Article Headlines, Article Sources, Company Name, Date of recorded price
  \item *5th version: Article Headlines, Article Sources, Company Name
  \item *6th version: Article Headlines, Article Sources, Company Name, Date of recorded price
\end{itemize}
\textit{* indicates the version was trained symbolically, given only 1 or -1 based on the sign of the actual value: 2.1232 is substituted for 1, -3.0832 is substituted for -1, and so on}

\section{Model}
We opted to use a variation of the BERT family of models known as Bert-Tiny\cite{bhargava2021generalization}\cite{DBLP:journals/corr/abs-1908-08962} for the backbone, because it is small and lightweight, with only 14.5M parameters.
The BERT architecture begins by tokenizing the input, converting words into specialized numbers in between 0 and 1. This tokenization process generated both word embeddings and input embeddings.
\begin{enumerate}
    \item Word Embeddings: each word is assigned a unique set of numbers, and every time that specific word is used, the same word embeddings will exist in that positional space in the word embeddings array.
    \item Positional embeddings: each word is placed in the context of the words surrounding it. This means that while the word "out" might be used in 2 parts of the sentence, it has a different context and potential meaning each time.
\end{enumerate}
Finally, both of these sets of embeddings are combined to form the input embeddings. 

Next, the input embeddings are passed through the BERT model itself: first, multi-head attention nodes, which contribute to the model being able to realize the context of each word. This aids analysis. Then, the input is passed through a neural network and activated using the GELU function. This process is repeated N = 2 times. 
A dropout layer is next applied. Then comes the loss function. Mean-Squared Error(MSE) loss is then applied.
\begin{equation}
MSE = \frac{1}{n} \sum_{i=1}^{n} (\hat{y}_i - y_i)^2
\end{equation}
The MSE loss algorithm works by squaring the difference between the predicted and real values, which serves two key purposes. The first is to make all error values positive, so the MSE error can emphasize the accurate magnitude of errors rather than their direction. The second purpose of the squaring function is to discriminate based on the magnitude of error: large error values will be that much larger when squared, while smaller error values will become less significant when squared. This also means large outliers, such as data points that vary drastically from the average, will be weighted heavily. MSE overall enables the model to focus on avoiding large, significant errors.
Finally, dropout, MSE loss, and the Adam optimization algorithm is applied before the prediction is completed. 
\begin{figure}[H]
    \centering
    \includegraphics[width=0.5\linewidth]{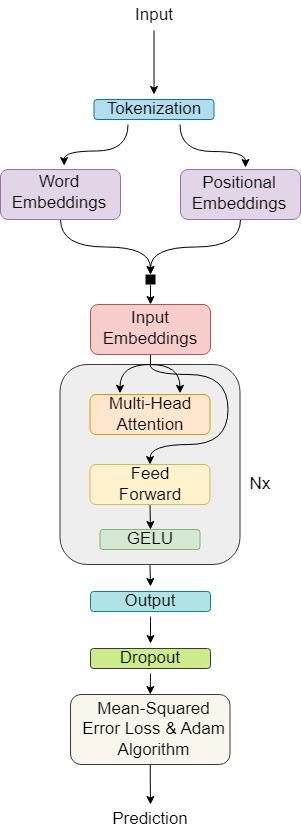}
    \caption{The BERT-Tiny Architecture}
    \label{fig:BERT-Tiny Architecture}
\end{figure}
\section{Experimental Results}

We analyzed our models in several different ways. First, we evaluated the models by to price direction accuracy.
We used mean squared error (MSE) accuracy as an evaluation metric to gauge the performance of each model. 
The testing data used to measure model performance was obtained by dividing our data into two groups: testing and training. We opted for a 10:90 split, meaning 10\% of our original data became the testing data while the remaining 90\% was used to train the model. Both the testing and training classes were created at random from the collected data.
\subsection{LSTM Comparison}
Our model groups compared to the LSTM standard model on testing data collected periodically over the span of 1 year are as follows:

\begin{figure}[H]
    \centering
    \includegraphics[width=1\linewidth]{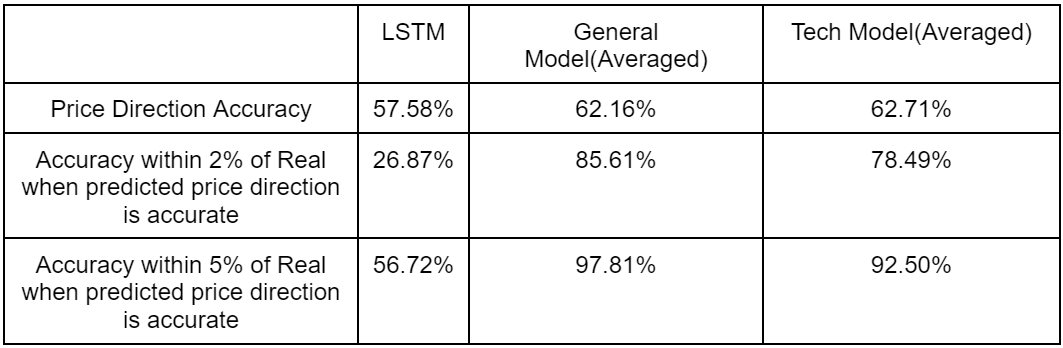}
    \caption{BERT vs LSTM Metrics}
    \label{fig:BERT vs LSTM Metrics}
\end{figure}

\textit{Note that the accuracy within 2\% and accuracy within 5\% when the prediction is in the proper direction is only measured with the first 4 versions of each BERT model class, since the last 2 versions of models were not trying to approach the exact model price but symbolically indicate direction, and weren't even trained on exact prices. Thus, we cannot expect our model to predict them, and any statistic including their approximation to the actual price would be inaccurate.}

As shown in the table above, the LSTM seems to perform just a bit worse than the general and tech models. The LSTM performed much worse than our models when rated on how close to the real price it predicted.

However, the statistics above measured the average accuracy of models trained using a variety of modalities. The higher performing models of each version compared differently to the LSTM. This demonstrates that not all models are superior to LSTM, but the average variant of our models exceeds the performance of the classic LSTM.

\begin{figure}[H]
    \centering
    \includegraphics[width=\linewidth]{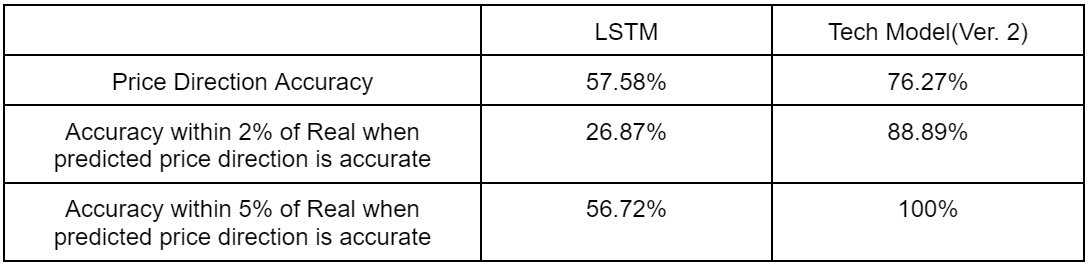}
    \caption{Tech Model (Ver. 2) vs. LSTM}
    \label{fig:Tech Model(Ver. 2) compared to LSTM}
\end{figure}
\subsection{Model Accuracy Statistics}
\begin{figure}[H]
    \centering
    \includegraphics[width=1\linewidth]{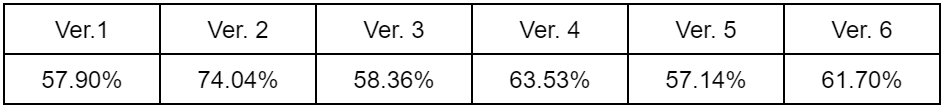}
    \caption{General Model Price Direction Accuracy}
    \label{fig:General Model Price Direction Accuracy}
\end{figure}
\begin{figure}[H]
    \centering
    \includegraphics[width=1\linewidth]{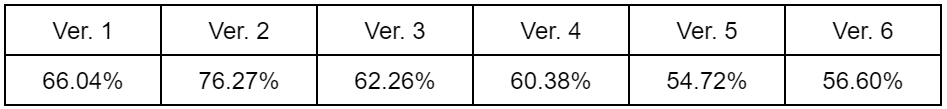}
    \caption{Tech Model Price Direction Accuracy}
    \label{fig:Tech Model Price Direction Accuracy}
\end{figure}

In addition, there are patterns represented in each version's accuracy that speaks to the efficacy of the modalities used in each version. 
One important trait of our model was its overall accuracy, in the sense that it is predicting not only individual prices, but a specific trend and pattern as a whole. As we use more and more validation data with the model, the more accurately it begins to predict the overall trend.
Using a larger scope, our model is generally able to accurately predict stock prices and stock trends when simply given a collection of news headlines.
\begin{figure}[H]
\centering
\includegraphics[width=5cm]{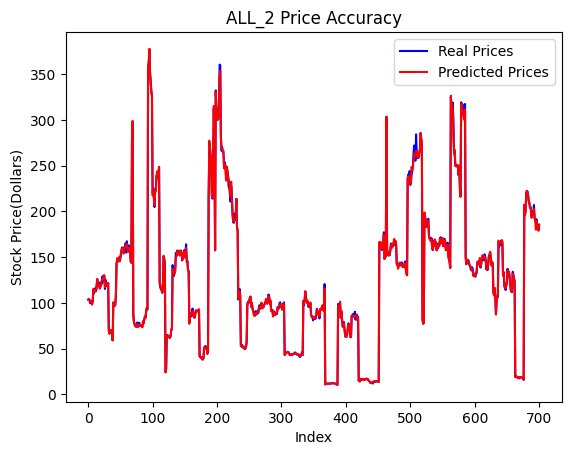}
\caption{Accuracy Graph of General 2 Model}
\label{fig:Accuracy Graph}
\end{figure}
\begin{figure}
    \centering
    \includegraphics[width=0.5\linewidth]{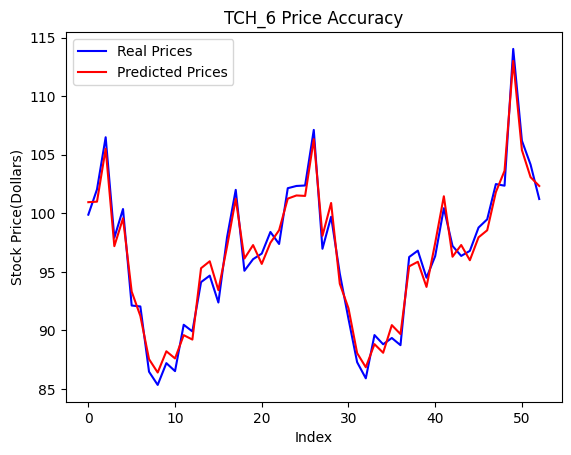}
    \caption{Accuracy Graph of Tech 6 Model}
    \label{fig:Accuracy Graph 6}
\end{figure}
As evident by the charts above, the models are capable of following long term trends proficiently. While short-term case-by-case instances, such as individual accuracy ratings, might not be as high (74.04\% and 56.66\% price direction accuracy respectively), as time continues, the model is able to adapt quite well to the trend of the data. 
The model's ability to adapt to long-term data trends is important, because while it may be only proficient in predicting small stock movements, the ability to track and predict overall trends is much more useful in long-term investment strategies than day-by-day cases.
\subsection{Comparison With Similar Models}
Our accuracy can be compared to a similar type of model on stock price prediction\cite{9434810}. It is important to note that this model was only tested on specific stocks, while our model was given a much more diverse pool of stocks as inputs. It is also worth noting that their model was tested on less data than ours was.
\begin{figure}[H]
    \centering
    \includegraphics[width=1\linewidth]{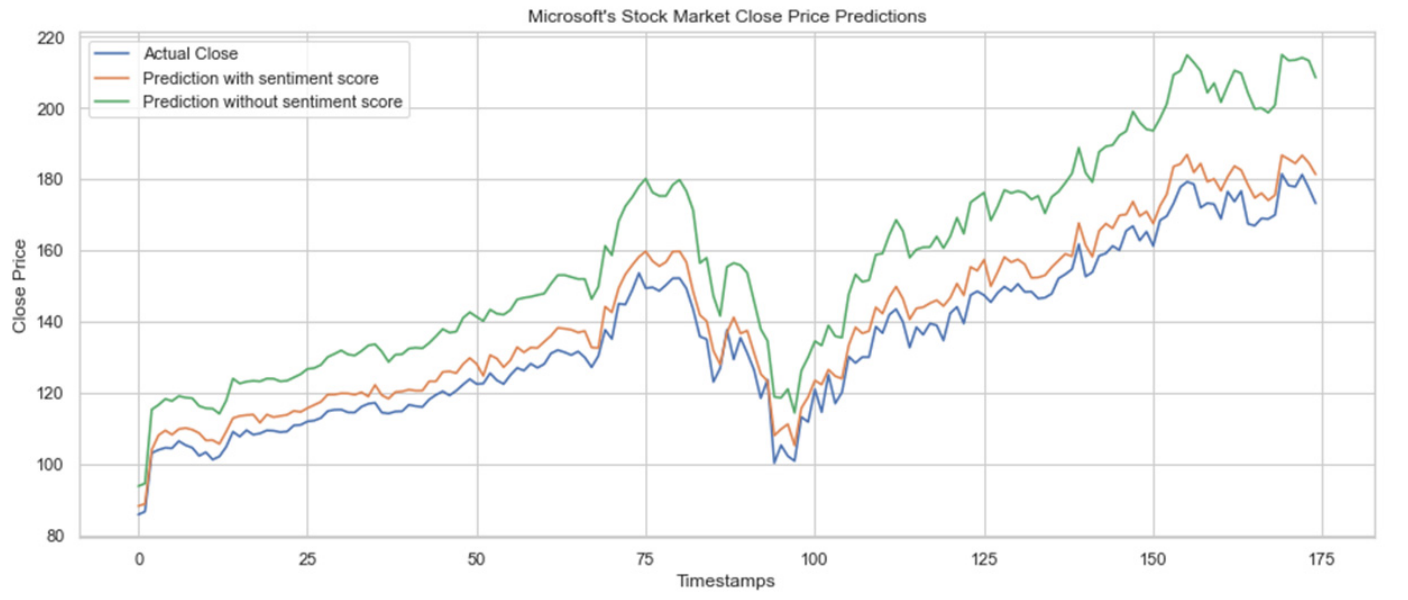}
    \caption{}
\end{figure}
\begin{figure}[H]
    \centering
    \includegraphics[width=1\linewidth]{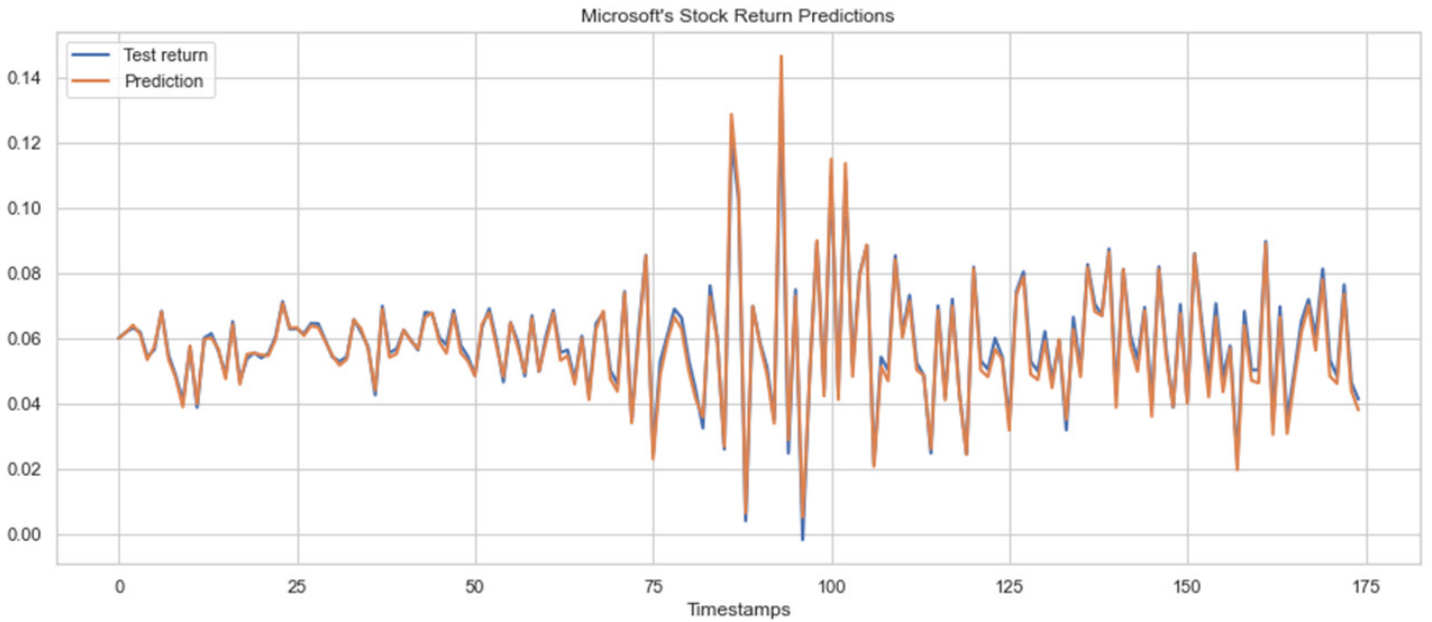}
    \caption{}
\end{figure}
As visible in our model accuracy graphs and the graphs of similar models displayed above, both of which relatively match the real stock price with their predictions, our model's accuracy over time is quite comparable to models proposed by similar papers.

\section{Discussion and Analysis}
The BERT models we have developed have thus far shown promise in predicting the stock market, especially in terms of overall trends, but they also introduce several illuminating concepts about the process of stock prediction itself. One such concept is observed in the metrics of models of version 2. In both types of model, general and tech-based, the 2nd version was always the most successful at predicting stock direction and was also the most accurate at predicting precise price, as version 2 models were also very accurate in comparison with models of other versions.
\begin{figure}[H]
    \centering
    \includegraphics[width=1\linewidth]{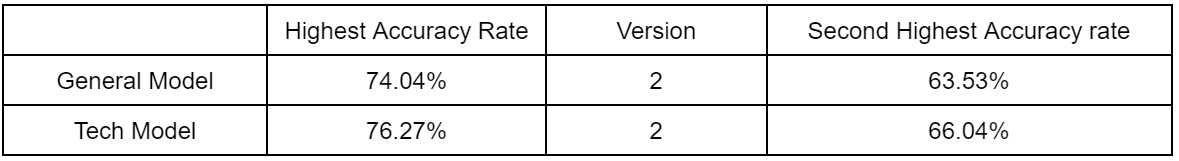}
    \caption{Version 2 Models Comparison}
    \label{fig:Version 2 Models Comparison}
\end{figure}

This version was only fed information about the company name and article headline. It was restricted from the date and source of said article. It might intuitively make sense for the model to perform strongly here because the name of the target company and sentiment to be analyzed are both very strong indicators of stock performance. This isn't to say other information such as company name and date isn't as important(model version 4 performed well, see the general and tech model accuracy charts), but feeding only the most important data to the model might play a role in its accuracy. There might also be a bit of confusion with the model, as it attempts to find a relationship between more and more variables.

This observation illustrates a few key points. Firstly, it doesn't appear to matter where the article came from or by whom it was written. Adding this information serves to decrease accuracy, as evidenced by the significantly lower accuracy rates of models incorporating this information. Secondly, the date appears to have no effect on the algorithm. In fact, accuracy rates between identical models apart from access to date information perform very similarly. This might be because a factor such as the exact time of an article is just noise to the model; not every date and time is going to be significant because there isn't always a correlation between the date an article is published and the price of a stock.
\begin{figure}[H]
    \centering
    \includegraphics[width=1\linewidth]{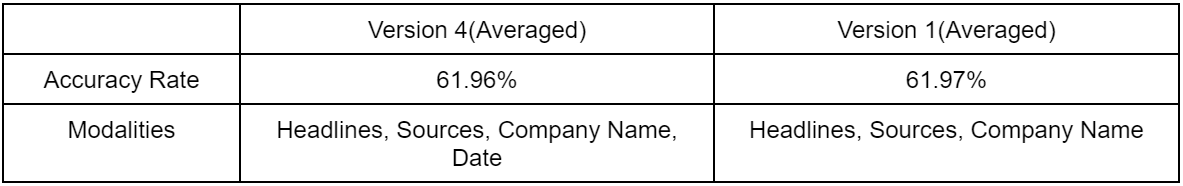}
    \caption{Date Information Effect on Model Accuracy}
    \label{fig:Date Information Effect on Model Accuracy}
\end{figure}

Of course, as mentioned earlier, all of these prices were collected in the window of the 2022-2023 year, so their accuracy is only known during this time period. This does give the models the added benefit of not being influenced by biased data such as that during the early 2000s, in which the technology market was very different due to the dot com bubble in part.

Another trend in the metrics of our models is the comparison between the symbolically trained models and the non-symbolically trained models. For example, both versions 1 and 5 are identical save for the fact that version 5 models are symbolic, and version 1 models are not. The same goes for models of version 4 and, it's symbolic clone, version 6. When we examine the accuracy disparity between the two, we see that there is not a significant shift in accuracy between the symbolic and regular models. The regular models outperform the symbolic models by some degree, but this difference  in performance ranging from a fraction of a percent to just under 4\%. This proves that symbolic training doesn't have a significant impact on the performance of a model. In fact, looking at the minute details, it is more beneficial to not train symbolically, as the models that tended to train based on exact stock prices ended up scoring higher in directional accuracy than those that trained symbolically. This indicates that having exact price data may be more beneficial in some cases than training based on price direction alone.

\begin{figure} [H]
    \centering
    \includegraphics[width=\linewidth]{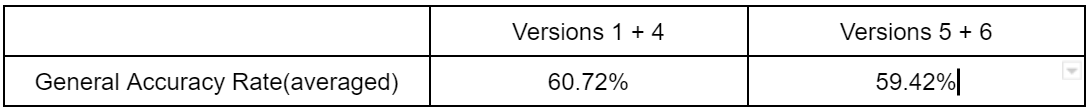}
    \caption{Symbolic v Non-Symbolic}
    \label{fig:Symbolic v Non-Symbolic}
\end{figure}

Finally, there is the question of grouping. Does a model trained on sector-specific stocks perform better because it is able to grasp certain characteristics of the field? For example, tech stocks tend to be more volatile when compared to beverage stocks. The metrics tell us it might not matter if models are trained on sector-specific datasets. It is true that the model with the highest accuracy rate belongs to the sector-specific tech-based group, but so does the model with the lowest accuracy rate. This indicates that specific stocks don't necessarily need to be grouped with similar ones in order to be accurately predicted, that a model trained on a variety of stocks can still function just as well.

\section{Conclusion}
While this paper has been able to explore avenues such as useful modalities and strategies for stock price prediction, there is still much more to be uncovered. 
The model can't accurately predict every stock trend, or infinitely into the future, but it might indicate general, high-level trends, which can be useful for investors as a sort of aid. Investors can work without models like this, and models like this can work without investors, but a combination of human thought and machine learning might be able to strike a certain balance and potentially outperform sole humans or models. For example, using models such as ours that are generally accurate can be useful to help human investors make decisions about whether to buy or sell a stock, but such models don't always necessarily need to determine the overall action.
Further research might include a further in-depth analysis of what specific modalities are necessary to accurately predict the price of a stock. This can be found via more extensive trial-and-error, with the addition of modalities such as average sector loss/gain on a given date, total loss/gain for that particular stock on a given date, and cost of a given stock relative to the average of stocks in the same sector. Further studies may also explore integrating a raw price factor into the algorithm in order to see the effect of that information.

\bibliographystyle{IEEEtran}
\bibliography{references}

\end{document}